\documentclass[12pt]{article}

\usepackage[english]{babel}

\usepackage[a4paper,top=2cm,bottom=2cm,left=3cm,right=3cm,marginparwidth=1.75cm]{geometry}

\usepackage{comment}
\usepackage[dvipsnames]{xcolor}
\usepackage{amsmath}
\usepackage{graphicx}
\usepackage[colorlinks=true, allcolors=blue]{hyperref}

\usepackage{bm}
\usepackage{algorithm}
\usepackage[noend]{algpseudocode}
\usepackage{multirow}
\usepackage{adjustbox}
\usepackage{colortbl}
\usepackage{array}
\usepackage{algorithm,algpseudocode,float}

\title{Does differentially private synthetic data lead to synthetic discoveries?}
\author{Ileana Montoya Perez \and Parisa Movahedi \and Valtteri Nieminen \and Antti Airola \and Tapio Pahikkala}
\date{Department of Computing, University of Turku, Finland}
\begin{document}
\maketitle

\begin{abstract}
Background: Synthetic data has been proposed as a solution for sharing anonymized versions of sensitive biomedical datasets. Ideally, synthetic data should preserve the structure and statistical properties of the original data, while protecting the privacy of the individual subjects. Differential privacy (DP) is currently considered the gold standard approach for balancing this trade-off.  

Objectives: We investigate how trustworthy are group differences discovered by independent sample tests from DP-synthetic data. The evaluation is carried out in terms of the tests' Type I and Type II errors. With the former, we can quantify the tests' validity i.e. whether the probability of false discoveries is indeed below the significance level, and the latter indicates the tests' power in making real discoveries.

Methods: We evaluate the Mann-Whitney U test, Student's t-test, chi-squared test and median test on DP-synthetic data. The private synthetic datasets are generated from real-world data, including a prostate cancer dataset (n=500) and a cardiovascular dataset (n=70 000), as well as on bivariate and multivariate simulated data. Five different DP-synthetic data generation methods are evaluated, including two basic DP histogram release methods and MWEM, Private-PGM, and DP GAN algorithms.

Conclusion: A large portion of the evaluation results expressed dramatically inflated Type I errors, especially at privacy budget levels of $\epsilon\leq 1$. This result calls for caution when releasing and analyzing DP-synthetic data: low p-values may be obtained in statistical tests simply as a byproduct of the noise added to protect privacy. A DP smoothed histogram-based synthetic data generation method was shown to produce valid Type I error for all privacy levels tested but required a large original dataset size and a modest privacy budget ($\epsilon\geq 5$) in order to have reasonable Type II error levels.

\end{abstract}

%Keywords: differential privacy, synthetic data, hypothesis testing, statistical inference, Mann-Whitney U test 

\section{Introduction}

As the amount of health and medical data collected from individuals has grown, so has the interest in using it for secondary purposes such as research and innovation. Many benefits have been proposed to arise from sharing this data \cite{el2015anonymising}, for example, enhancing research reproducibility, building on existing research, performing meta-analyses, and reducing clinical trial costs by reusing existing data. However, privacy concerns about the potential harm to individuals that may come from sharing their sensitive data, along with legislation aimed at addressing these concerns, restrict the opportunities for sharing individuals' data.

The release of synthetic data, generated using a statistical model derived from an original sensitive dataset, has been proposed as a potential solution for sharing biomedical data while preserving individuals' privacy \cite{rubin1993statistical,chen2021synthetic,hernadez2023synthetic}. It has been argued that since synthetic data consists of synthetic records instead of actual records, and synthetic records are not associated with any specific identity, privacy is preserved \cite{rubin1993statistical}. However, it has been repeatedly demonstrated that this is not the case as synthetic data is not inherently privacy-preserving \cite{jordon2022synthetic_turing, gan_leaks_chen2020, logan_membership_inference_attacks_on_gan, stadler2022synthetic}. In the worst case, a generative model could create near copies of the original sensitive data it was trained on. Moreover, there are many more subtle ways that models can leak information about their training data \cite{jordon2022synthetic_turing, carlini2019secret}. At the other extreme, perfect anonymity is guaranteed only when no useful information from the original data remains. Therefore, in addition to preserving privacy, the generated data should have high utility, meaning the degree to which the inferences obtained from the synthetic data correspond to inferences obtained from the original data \cite{jordon2022synthetic_turing, karr2006framework_fidelity_utility}. Consequently, when generating synthetic data, it is essential to find a balance between the privacy and utility of the data, ensuring that the generated data captures the primary statistical trends in the original data while also preventing the disclosure of sensitive information about individuals \cite{boedihardjo2022covariance}.

Differential privacy (DP), a mathematical formulation that provides probabilistic guarantees on the privacy risk associated with disclosing the output of a computational task, has been widely accepted as the gold standard of privacy protection \cite{dwork2006calibrating,dwork2014algorithmic,Wasserman2010,Gong2020}. As a result, methods that ensure DP guarantees have been introduced in a broad range of settings, including descriptive statistics \cite{dwork2014algorithmic, xu2013differentially}, inferential statistics \cite{gaboardi2016differentially, task2016differentially, couch2019differentially, ferrando2022parametric}, and machine learning applications \cite{Gong2020, chaudhuri2011differentially}. Furthermore, DP offers a theoretically well-founded approach that provides probabilistic privacy guarantees also for the release of synthetic data. Therefore, several methods for releasing DP-synthetic data have been proposed (see e.g. \cite{hardt2012simple, ping2017datasynthesizer, snoke2018pmse, chen2020gs, McKenna2021}). Some state-of-the-art methods for generating DP-synthetic data use multi-dimensional histograms, which are standard tools for estimating the distribution of data with minimal a priori assumptions about its statistical properties. Other methods are based on machine learning-based generative models, for example, Bayesian and Generative Adversarial Network (GAN) based methods. The aim of DP-synthetic data is to be a privacy-preserving version of the original data that could be safely used in its place, requiring no expertise on DP or changes to the workflow from the end-user. However, DP-synthetic data is always a distorted version of the original data, and especially when high levels of privacy are enforced the level of distortion can be quite considerable. Even though combining DP with synthetic data guarantees a desired level of privacy, preservation of the utility remains unclear. In particular, the validity of statistical significance tests, namely the statistical guarantees of the false finding probabilities being at most the significance level, may be lost. 

Hypothesis tests for assessing whether two distributions share a certain property are essential tools in analyzing biomedical data. In this work we particularly focus on the Mann-Whitney (MW) U test (a.k.a. Wilcoxon rank-sum test or Mann-Whitney-Wilcoxon test), as it is the de facto standard for testing whether two groups are drawn from the same distribution \cite{nachar2008mann, zar2010biostatistical}. It is widely applied in medical research \cite{okeh2009statistical}, particularly when analyzing a biomarker between non-healthy and healthy patients in clinical trials. It is well known (see e.g. \cite{fay2010wilcoxon}) that MW U test is valid for this question, that is, the probability of falsely rejecting the null hypothesis of the two groups being drawn from the same distribution is at most the significance level determined a priori. Alongside the MW U test, we also consider the Student’s t-test \cite{kim2015t}, median test \cite{conover1999practical}, and chi-squared test \cite{mchugh2013chi}. In general, the choice of statistical test should be guided by the distribution characteristics of the dataset and type of variable under analysis.

In order for DP-synthetic data to be useful for basic use cases in medical research, such as the MW U test, one would hope to observe roughly similar results when carrying out tests on sensitive medical datasets. Otherwise, there is a risk that discoveries are missed because of information lost in synthetization, or worse, that false discoveries are made due to artifacts introduced in the data generation process.

\section{Objectives}

\begin{figure} [ht]
    \centering
     \includegraphics[scale=0.65]{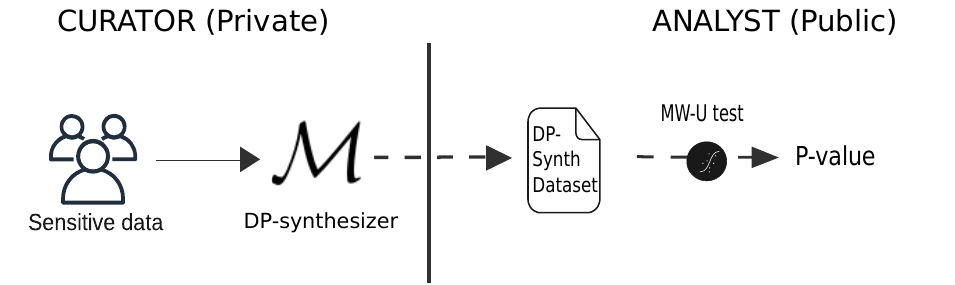}
    \caption{\small The overall configuration of the study.
    }
    \label{fig:figure_1}
\end{figure}

DP-synthetic data has been proposed as a solution for publicly releasing anonymized versions of sensitive data such as medical records. Ideally, this would allow for performing reliable statistical analyses on the DP-synthetic data without ever needing to access the original data (see Figure \ref{fig:figure_1}). However, there is a risk that DP-synthetic data generation methods distort the original data in ways that can lead to loss of information and even to false discoveries.

In this study, we empirically evaluate the validity and power of independent sample tests, such as the MW U test, applied to DP-synthetic data. The Type I and Type II errors are used to measure the test validity and power, respectively. On one hand, a test is valid if, for any significance level, the probability that it falsely rejects a correct null hypothesis is no larger than the significance level \cite{casella2002statistical}. If the test is not valid, its use can lead to false scientific discoveries, and hence its practical utility can be even negative. On the other hand, the test's power refers to the probability of correctly rejecting a false null hypothesis.

In our experiments with the MW U test, we evaluated five different DP-synthetic data generation methods on bivariate real-world medical datasets, as well as data drawn from two Gaussian distributions. Additionally, we performed experiments with simulated multivariate data to explore the behavior of MW U test, Student’s t-test, median test, and chi-squared test on higher-dimensional DP-synthetic data consisting of different variable types. Our study contributes to understanding the reliability of statistical analysis when DP-synthetic data is utilized as a proxy for private data whose public release is challenging or even impossible.

\section{Methods}
In this section, we first present the formal definition of DP.  Next, we introduce DP methods for synthetic data generation while describing the five DP-synthesizers used in this study. Following that, we explain the validity and power of a statistical test. Finally, we introduce the independent sample tests considered in this study. 

\subsection{Differential Privacy}

DP is a mathematical definition that makes it possible to quantify privacy \cite{dwork2006calibrating, dwork2014algorithmic}. A randomized algorithm $\mathcal{M}$ satisfies $(\epsilon,\delta)$-differential privacy if for all outputs $S$ of $\mathcal{M}$ and for all possible neighboring datasets $D,D'$ that differ by only one row,
\begin{equation}\label{dpeq} 
Pr[\mathcal{M}(D)=S]\leq e^{\epsilon}Pr[\mathcal{M}(D')=S]+\delta
\end{equation}
where $\epsilon$ is an upper bound on the privacy loss, and $\delta$ is a small constant corresponding to a small probability of breaking the DP constraints. For $\delta=0$ in particular, solving (\ref{dpeq}) wrt. $\epsilon$ results to:
\begin{align}\label{logprobs}
\log(Pr[\mathcal{M}(D)=S])-\log(Pr[\mathcal{M}(D')=S])\leq {\epsilon}
\end{align}
indicating that the log-probability of any output can change no more than $\epsilon$. Accordingly, an algorithm $\mathcal{M}$ which is $\epsilon$-DP guarantees that, for every run of $\mathcal{M}(D)$ the outcome obtained is almost equally likely to be
obtained on any neighboring dataset, bounded by the value of $\epsilon$. Informally, in DP, privacy is understood to be protected at a given level of $\epsilon$ if the algorithm's output does not overly depend on the input data of any single contributor; it should yield a similar result if the individual's information is present in the input or not. 

Typically, DP methods are non-private methods that are transformed to fulfill the DP definition. This is achieved by adding noise using a noise mechanism calibrated based on the $\epsilon$ and the algorithm to be privatized \cite{dwork2006calibrating, dwork2014algorithmic}. Choosing the appropriate value of epsilon is context-specific and an open question, but, for example, $\epsilon \leq 1$ has been proposed to provide a strong guarantee \cite{arnold2020really}, while $1 < \epsilon \leq 10$ is considered to still produce useful privacy guarantees \cite{abadi2016deep}, depending on the task and type of data.

\subsection{DP Methods for Synthetic Data Generation}

In recent years, several methods for generating DP-synthetic data have been proposed \cite{ping2017datasynthesizer, snoke2018pmse, McKenna2021, abay2019privacy, Jordon2019}. Some of the proposed methods are based on histograms or marginals. These methods privatize the cell counts or proportions of a cross-tabulation of the original sensitive data to generate the DP-synthetic data. Other methods use a parameterized distribution or a generative model that has been privately derived from the original data. While DP methods based on histograms or marginals have been found to produce usable DP-synthetic data with a reasonable level of privacy guarantee, methods based on parameterized distributions or deep learning-based generative models have presented greater challenges \cite{Bowen2019ComparativeSO, Bowen2016ComparativeSO}. 

Generative methods based on marginals share a three-step process: initially, a set of marginals is identified, either manually by a domain expert or through DP-automatic selection. Next, these chosen marginals are measured using DP. Finally, synthetic data is generated from the noisy marginals.
To address the challenges of high-dimensional domains, recent methods have been developed to automatically and privately select a subset of marginals ensuring their preservation in the synthetic data generated, such as PrivMRF \cite{cai2021data}, Privbayes \cite{10.1145/3134428}, MWEM \cite{hardt2012simple} and AIM \cite{mckenna2022aim}.

PrivMRF employs Markov Random Fields to generate synthetic data under differential privacy, emphasizing the retention of statistical correlations between selected marginals within the privacy constraints. PrivBayes constructs a Bayesian network under differential privacy, utilizing a selected set of marginals to approximate the underlying data distribution for synthetic data generation. The MWEM algorithm is designed to generate a data distribution that yields query responses closely resembling those obtained from the actual dataset. AIM on the other hand is a workload-adaptive algorithm, allowing for the input of a pre-defined set of marginals to be specifically preserved in the final DP approximated distribution.

There are two approaches to consider when designing a DP workflow: global DP and local DP \cite{dwork2014algorithmic}. Global DP  involves aggregating data in a central location and is managed by a trusted curator, ensuring privacy at the dataset level. In contrast, Local DP decentralizes the privacy mechanism by applying it directly to the individual's data before it is shared. Many applications, such as crowdsourced systems, involve data distributed across multiple individuals who do not trust any other party. These individuals are only willing to share their information if it has been privatized on their own devices prior to transmission. In such cases, local privacy methods such as LoPub and LoCop \cite{wang2019locally,ren2018textsf,chen2016private} become applicable, ensuring that each individual's data remains confidential even when aggregated from diverse sources.

In this study, we focus on five well-known DP methods for generating synthetic data in a global DP setting. These methods have established algorithms or available packages, making them accessible to any practitioner. Following, we provide a brief description of each of these DP methods.

\begin{itemize}
\item \textbf{DP Perturbed Histogram}

This method uses the Laplace mechanism \cite{dwork2014algorithmic} to privatize the original histogram bin counts. The noise added to each bin is sampled separately from a calibrated Laplace distribution. After adding the noise, all negative counts are set to zero, and individual-level data is generated from the noisy counts.

\item \textbf{DP Smoothed Histogram}

This method generates synthetic data by randomly sampling from the probability distribution determined by the following histogram. The probabilities of the histogram bins are proportional to $c_i+2m/\epsilon$, where $c_i$ is the number of original data points in the $i$th histogram bin and $m$ is the size of the synthetic dataset drawn. The approach is similar to the one discussed by Wasserman and Zhou \cite{Wasserman2010}. Unlike the other considered DP methods, the utility of this method is inversely proportional also to the amount of synthetic data drawn. Therefore, in our experiments, we use the method only in settings where the size of the synthetic data generated is considerably smaller than that of the original sensitive data. A proof of the approach being DP is presented in the supplementary material \ref{appendix:proof}.

\item \textbf{Multiplicative Weights Exponential Mechanism (MWEM)}

This algorithm proposed by Hardt et al.\cite{hardt2012simple} is based on a combination of the multiplicative weights update rule with the exponential mechanism. The MWEM algorithm estimates the original data distribution using a DP iterative process. Here, a uniform distribution over the variables of the original data is updated using the multiplicative weighting of a query or bin count selected through the exponential mechanism and privatized with the Laplace mechanism in each iteration. The privacy budget $\epsilon$ is split by the number of iterations, as in every iteration the original data needs to be accessed. 

\item \textbf{Private-PGM}

McKenna et al.\cite{McKenna2021} propose this approach for DP-synthetic data generation. It consists of three basic steps: 1) Selecting a set of low-dimensional marginals (i.e., queries) from the original data. 2) Adding calibrated noise to the marginals. 3) Generating synthetic data that best explains the noisy marginals. In step 3, based on the noisy marginals, a probabilistic graphical model (PGM) is used to determine the data distribution that best captures the variables' relationship and enables synthetic data generation.

\item \textbf{Differentially private GAN (DP GAN)}

Generative adversarial networks (GAN) \cite{goodfellow2020generative} consist of a generator, denoted with \textbf{G}, and one or more discriminators \textbf{D}. The goal is, that \textbf{G} would learn to produce synthetic data similar to the original data. The two networks are initialized randomly and trained iteratively in a game-like setup: \textbf{G} is fed noise to create synthetic data, which the \textbf{D} tries to discriminate as being original or synthetic. The generator uses feedback from the discriminator(s) to update its parameters via gradient descent (for a detailed explanation see \cite{goodfellow2016nips}). GANs, and other deep learning models, can attain privacy guarantees by using a DP version of an optimization algorithm, most often differentially private stochastic gradient descent (DPSGD) \cite{abadi2016deep}.

\end{itemize}

\subsection{Validity and Power of Independent Sample Tests}

Samples are considered independent when individuals in one group do not influence or share information with individuals in another group. Each group consists of unique members, and no pairing or matching occurs between them. To evaluate potential statistical differences between the two groups, researchers commonly use statistical tests designed for independent samples. These tests determine whether the samples were drawn independently from distributions with shared properties. The independent sample tests considered in this work are the MW U test, Student’s t-test, median test, and chi-squared test.

 The validity and power of a statistical test can be evaluated in terms of Type I and Type II errors. Let us recall that Type I is the error incurred when a “True” null hypothesis is rejected, producing false inference. On the other hand, Type II is the error of failing to reject a “False” null hypothesis (see Figure \ref{fig:figure_2}). Following \cite{casella2002statistical}, we say that p-value corresponding to the observed test statistic is valid, if it is at most the probability of observing as extreme test statistic under the null hypothesis. Consequently, the significance test is valid if its p-value is valid.  

 A priori selected significance level $\alpha$ defines a threshold that, for any valid hypothesis test, forms an upper bound on the probability of committing Type I error. 
A typical choice for $\alpha$ is 0.05, indicating a maximum 5\% chance of incorrectly rejecting a true null hypothesis. The probability of making a Type II error is often denoted as $\beta$ (beta), from which the power of the test can be determined by computing $1-\beta$. The power of a test can be interpreted as the probability of correctly rejecting a null hypothesis when it is in fact "False". The power depends on the analysis task, being affected by factors such as chosen significance level, the effect size, the sample size, and the relative sizes of the different groups. In our experiments, we observed the imbalance between group sizes to have a dominant effect on tests' power in practice, because of the DP-synthetic data generators' tendency to produce imbalanced samples for small $\epsilon$ values.

\subsection{Mann-Whitney U Test}

The MW U test is a statistical test first proposed by Frank Wilcoxon \cite{wilcoxon1945} in 1945 and later, in 1947, formalized by Henry Mann and Donald Whitney \cite{mann1947test}. While there are many different uses and interpretations of the test (see e.g. \cite{fay2010wilcoxon} for a comprehensive review), in this paper we focus on the null hypothesis that two samples or groups are drawn from the same distribution. The test carried out on two groups produces a value of the MW U statistic and the corresponding p-value. The U statistic measures the difference between the groups as the number of times an observed member of the first group is smaller than that of the second group, ties being counted as a half time.
The p-value indicates the strength of evidence the value of the U statistic provides against the null hypothesis, given that the assumption of the data being independently drawn holds.

Couch et al. \cite{couch2019differentially} proposed a differentially private version of the Mann-Whitney U test (DP-MW). The DP-MW U test is presented as ($\epsilon$, $\delta$)-DP, where a portion of the privacy budget $\epsilon$, and $\delta$ are used for privatizing the smallest group size. The privatized size and the rest of $\epsilon$ are then used for privatizing the U statistic using a calibrated Laplace distribution. In order to calculate the corresponding p-value, the DP-MW U distribution under the null hypothesis is generated based on the privatized group sizes. Detailed information and algorithms are provided by Couch et al. \cite{couch2019differentially}.
The DP-MW U test is not based on analyzing synthetic data, but rather the test is carried out directly on the original sensitive dataset, and DP guarantees that sensitive information about individuals is not leaked when releasing the test results. 
 
In this study, the DP-MW U test on the original sensitive data provides us with a reference point, a valid test with the best-known achievable power when performing MW U test under DP. In contrast, the ordinary MW U test is evaluated on the DP-synthetic data. If the validity of the ordinary test is preserved, comparison to the reference point indicates how much power is lost when general purpose DP-synthetic data is generated as an intermediate step.

\subsection{Student’s t-Test, Chi-squared Test and Median Test}
The Student's t-test (independent or unpaired t-test) \cite{kim2015t} is a widely utilized parametric statistical test that assesses whether the means of two independent samples are significantly different. The null hypothesis states that the means are statistically equivalent, while the alternative hypothesis suggests that they are not. The test is valid for two independent samples if their distributions are normal and their variances are equal.

The chi-squared test \cite{mchugh2013chi}, is a non-parametric test used to analyze the association of two categorical variables by utilizing a contingency table. Under the null hypothesis, the observed (joint) frequencies should equal to expected (marginal) frequencies, meaning that the variables are independent. Since, under the null hypothesis, the test statistic approximately follows a Chi-squared distribution, the test's validity depends on the sample size. However, for small $n$ and for $2\times2$ tables, the appropriate alternative is the Fisher's exact test.

Median test \cite{conover1999practical} is a nonparametric method used to test the null hypothesis of two (or more) independent samples being drawn from distributions of equal medians. The test is valid as long as the distributions have equal densities in the neighborhood of the median (see e.g. \cite{Freidlin2000median} and references therein).

\begin{figure}
    \centering
     \includegraphics[scale=0.35]{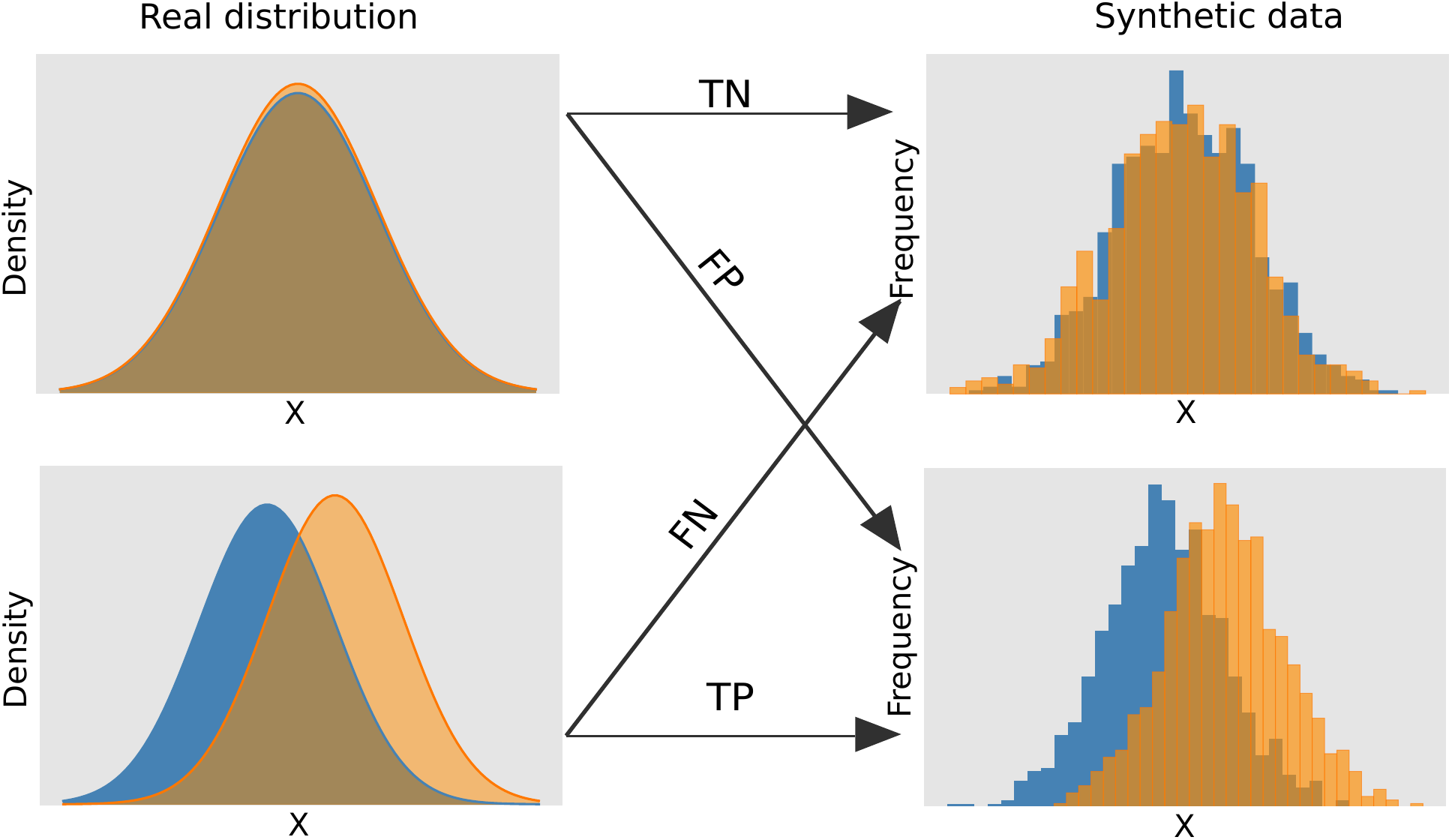}
    \caption{\small Possible outcomes of a hypothesis test that tests whether two distributions are the same. TN: true negative, TP: true positive, FP: false positive (Type I error), FN: false negative (Type II error).}
    \label{fig:figure_2}
\end{figure}

\section{Experimental Evaluation}
To empirically evaluate the utility of independent sample tests applied to DP-synthetic datasets, we conducted a set of experiments. In each experiment, either simulated or real-world data was used to represent the original sensitive dataset. This data was subsequently used to train DP-synthetic data generation methods. Finally, the independent sample tests were carried out on synthetic data produced by the generator.

First, we examined the  behavior of MW U test on DP-synthetic data generated based on bivariate real-world datasets or simulated data drawn from Gaussian distributions. As depicted by real distribution in Figure~\ref{fig:figure_2}, we considered two cases for Gaussian data: one where both groups are drawn from the same distribution (i.e., the null hypothesis is true) and one where they are drawn from distributions with different means (i.e., the null hypothesis is false). While in practice synthesizing datasets consisting of only two variables would have quite limited use cases, these experiments allow demonstrating the fundamentals of how different DP synthetization approaches affect the validity and power of  statistical tests.
In order to provide a more realistic setup, we further performed experiments on a simulated multivariate dataset. The validity and power of the MW U test, Student’s t-test, median test, and chi-squared test were explored in these experiments.

In the overall study design (see Figure~\ref{fig:figure_1}), the real-world, Gaussian, and simulated multivariate datasets correspond to the sensitive data given as input to a DP-synthesizer method that produces a DP-synthetic dataset as output.
In the following subsections, we present the datasets, the implementation details of the DP-synthetic data generation methods used, and the experiments conducted.

\subsection{Original Datasets}
First, we experimented with a setup, where the sensitive original data consists of only two variables (i.e., a binary variable and a continuous variable). The binary variable represents group membership (e.g. healthy or non-healthy), while the continuous variable is the one used to compare the groups with the MW U test.

To establish a controlled environment where the amount of signal (i.e., the effect size) in the population is known, we drew two groups of data from two Gaussian distributions with a known mean ($\mu$) and standard deviation ($\sigma$). More precisely, for non-signal data, which corresponds to a setting where the null hypothesis is true, the two groups were randomly drawn from the same Gaussian distribution ($\mu=50, \sigma=2$). For the signal data, which corresponds to a setting where the null hypothesis is false, two Gaussian distributions with effect size $\mu_1-\mu_2 = \sigma$ (i.e., $\mu_1 = 51, \sigma_1=1, \mu_2 = 50, \sigma_2=1$) were used to sample each group. Additionally, for those DP methods based on histograms or marginals, the sampled values for each group were discretized into 100 bins (ranging from 1 to 100).

\begin{figure}[ht]
    \centering
     \includegraphics[scale=0.5]{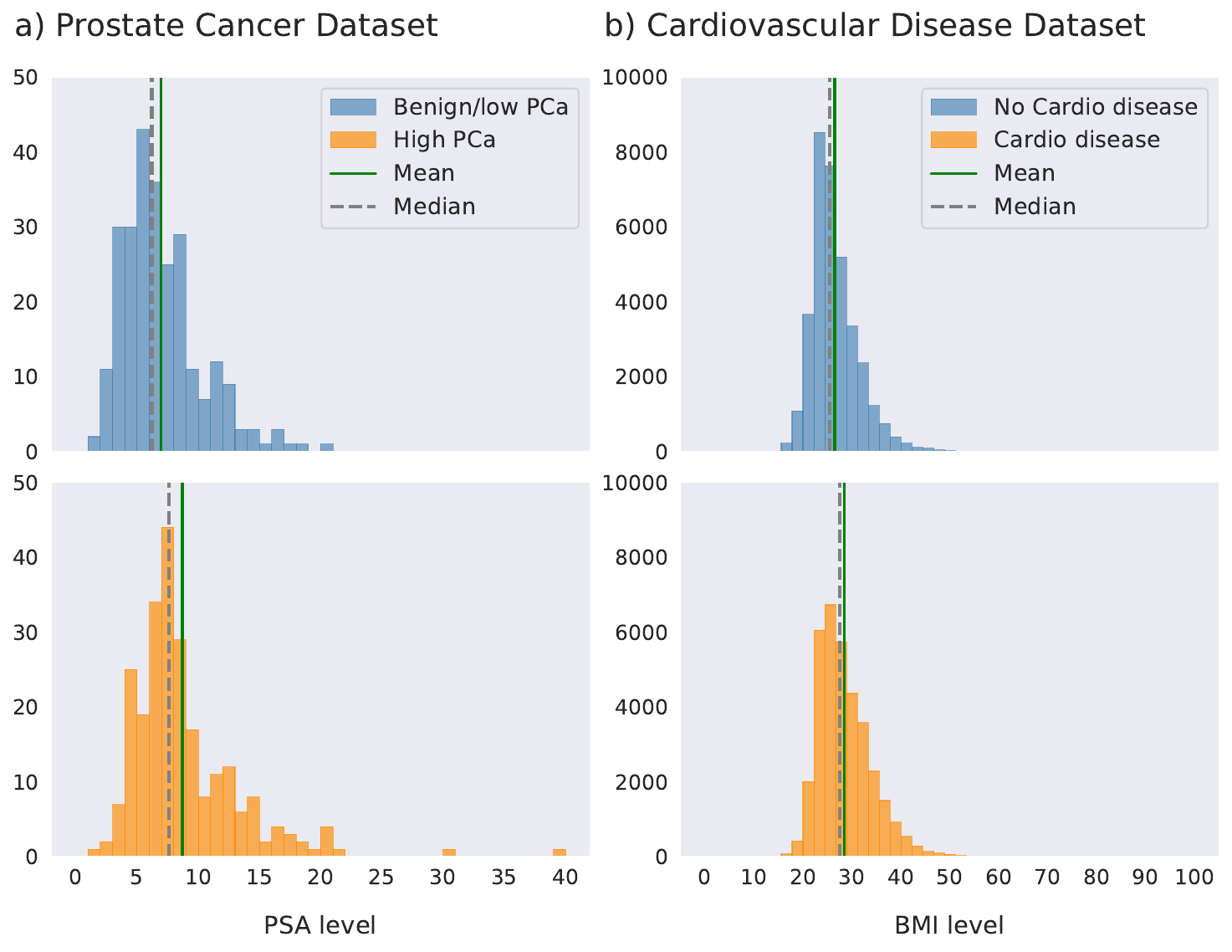}
    \caption{\small a) Prostate cancer (PCa) dataset: prostate-specific antigen (PSA) level distribution for high-risk and benign/low PCa. The difference between the groups is statistically significant (MW U stat= 22713, p-value= 1.4e-07), b) Kaggle Cardiovascular disease dataset: body mass index (BMI) distribution for subjects with the absence and presence of cardio disease. The difference between the groups is statistically significant (MW U stat= 471500929.50, p-value $\cong$ 0.000).
    }
    \label{fig:figure_3}
\end{figure}

In order to verify our experiment’s results on the Gaussian data, we also carried out experiments using real-world medical data. In this case, we use the following two datasets: 

\begin{itemize}
\item \textbf{The Prostate Cancer Dataset}

The data is from two registered clinical trials, IMPROD \cite{jambor2017novel} and MULTI-IMPROD \cite{jambor2019validation}, with trial numbers NCT01864135 and NCT02241122, respectively. These trials were approved by the Institutional Review Board, and each enrolled patient gave written informed consent. The dataset consists of 500 prostate cancer (PCa) patients (242 high-risk and 258 benign/low-risk PCa) with clinical variables, blood biomarkers, and MRI features. For our experiments, we selected two variables: a binary label that indicates the condition of the patient and the prostate-specific antigen (PSA) level. The PSA is a continuous variable that has been associated with the presence of PCa \cite{stamey1987prostate, catalona1991measurement}. Therefore, in this study, we considered the null hypothesis under test to be \textit{“The PSA level of high-risk and benign/low-risk PCa patients originate from the same distribution.”} Figure \ref{fig:figure_3}(a) presents the PSA distribution for both groups in this dataset. In those DP methods based on histograms or marginals, the PSA values were discretized using a 40 bins histogram (ranging from 1 to 40, where PSA $\geq$ 40 are in the last bin).

\item \textbf{Kaggle Cardiovascular Disease Dataset}

This dataset is publicly available \cite{Ulianova2019} and consists of 70 000 subjects and 12 variables, where the target variable is the cardio condition of the subjects, with 34 979 presenting cardiovascular disease and 35 021 without the disease. For our experiments, we use each subject body mass index (BMI), calculated from their weight and height, which has been related to cardiovascular conditions \cite{Larsson2020}. Here, the null hypothesis under test is \textit{“The BMI level for individuals with the presence of cardiovascular disease and the ones with absence cardiovascular disease originate from the same distribution.”} Figure \ref{fig:figure_3}(b) presents the distribution of both groups (i.e., cardio disease vs. no cardio disease). The BMI variable was discretized into 24 bins, where the first bin contains BMI $<$ 18 and the last bin BMI $\geq$ 40, in those DP methods that require it.

\end{itemize}

Finally, we experimented with simulated multivariate datasets. The simulation was based on the real-world prostate cancer dataset. The included variables were the patient’s age, PSA level, prostate volume, the use of 5-alpha-reductase inhibitors (5-ARIs) medication, prostate imaging reporting and data systems (PIRADS) score, and a class label indicating low-risk or high-risk PCa. The simulated datasets were generated by a GaussianCopulaSynthesizer from the Synthetic Data Vault (SDV) \cite{SDV} trained on the real-world dataset.
In the SVD settings, the age variable was configured to follow a normal distribution, while the remaining numerical variables were set to follow a beta distribution. In experiments with a false null hypothesis, SVD was conditioned to generate simulated datasets with an equal number of high-risk and low-risk patients. For experiments with a true null hypothesis, the condition was to generate only one class (low-risk) for the simulated dataset, and subsequently, half of the data were randomly assigned to the high-risk class.

\subsection{Implementations}
In our experiments, for the generated DP-synthetic data, we used the hypothesis tests provided by the Scipy v1.6.3 package \cite{2020SciPy-NMeth}, such as the \textit{mannwhitneyu} function for the MW U test. We used two-sided tests, with
all the tests statistics and p-values computed using the Scipy function's default values. As a point of reference, we also computed the DP-MW U statistic and p-value on the corresponding original sensitive dataset. The DP-MW U test was implemented using Python v3.7 and following the algorithms presented in \cite{couch2019differentially}, where 65\% of $\epsilon$ and $\delta = 10^{-6}$ are used for estimating the size of the smallest group, and the U statistic is privatized using the estimated size and the remaining $\epsilon$. 

In the case of the DP Perturbed Histogram, Python v3.7 was also used in the implementation. The noise, added to the original histogram, was sampled from a discrete Laplacian distribution \cite{Canonne2022} scaled by $\mathrm{\frac{2}{\epsilon}}$, then the noisy counts were normalized by the original dataset size. After that, the synthetic data was obtained by transforming the histogram counts to values using the bin center point. For Private-PGM \cite{McKenna_Private-PGM_2021} and MWEM \cite{Hardt2020}, their corresponding open-source packages were used to generate DP-synthetic data. The Private-PGM synthetic data was generated by following the demonstration in Python code presented by McKenna et al. \cite{McKenna2021} using Laplace distribution scaled by $\frac{2}{split(\epsilon)}$ where $split(\epsilon)$ is the privacy budget ($\epsilon$) divided by the number of marginal queries selected. MWEM was run with default hyperparameters; only $\epsilon$ was changed to show the effect of different privacy budgets. The resulting DP-synthetic data was sampled using the histogram noisy weights returned by the MWEM algorithm. The implementation of DP Smoothed Histogram was also coded in Python v.3.7 following Algorithm 1 provided in the supplement material. In all our experiments, the DP-synthetic data generators were configured to preserve all the one-way and two-way marginals.

The GAN model used is based on the GS-WGAN by Chen et al. \cite{chen2020gs}. The implementation is a modification of the freely available source code \cite{GSWGAN_github}, with changes made to suit tabular data generation instead of images. The generator architecture was changed from a convolutional- to a fully-connected three-layer network, and the gradient perturbation procedure was modified to accommodate these changes along with making the source code compatible with an up-to-date version of PyTorch (v1.10.2) \cite{paszke2019pytorch}. Hyperparameter settings were chosen based on the recommendations of Gulrajani et al. on the WGAN-GP \cite{gulrajani2017improved}, which of the GS-WGAN is a DP extension. This model uses privacy amplification by subsampling \cite{chen2020gs}, a strategy to achieve stronger privacy guarantees by splitting training data into mutually exclusive subsets according to a subsampling rate $\gamma$. Each subset is used to train one discriminator and the generator randomly queries one discriminator for one update.

\subsection{Experimental Setup}
In the experiments, we investigated the utility of the statistical test at different levels of privacy $\epsilon$. For the DP-MW U test and all DP-synthetic data generation methods, except for DP GAN, we used $\epsilon$ values of 0.01, 0.1, 1, 5, and 10. For the DP GAN experiments, the $\epsilon$ values were 1, 2, 3, 4, 5, and 10. The higher minimum of $\epsilon = 1$ was set due to differences between the DP GAN and the other methods. Every experiment was repeated 1000 times, and the proportion of Type I and Type II errors were computed and evaluated at a significance level $\alpha = 0.05$. 

\subsection*{\textbf{\small Setup for Gaussian Data}}
In our experiments on Gaussian data using the DP-MW U test, DP Perturbed Histogram, Private-PGM, and MWEM, each method was applied to original dataset sizes of 50, 100, 500, 1000, and 20 000 with a group ratio of 50\% and at the different values of $\epsilon$. In these methods, the original dataset size was considered to be of public knowledge, thus, the size of the generated DP-synthetic dataset was around or equal to the original size. 

Experiments with DP Smoothed Histogram were performed by randomly sampling original Gaussian dataset of large size (i.e., dataset size of 20 000 with a group ratio of 50\%). Then, the method was applied using the different values of $\epsilon$, and for every $\epsilon$ synthetic data of size 50, 100, 500, and 1000 were generated using the noisy probabilities returned by the method.

In all experiments with the GAN discriminator networks a subsampling rate $\gamma$, of 1/500 was used, resulting in mutually exclusive subsets of size 40. The sample size for the GAN training data was 20 000 in all settings and 1000 different generators were trained with models saved at the chosen values of $\epsilon$ (1, 2,3, 4, 5, and 10). Five synthetic datasets of sizes 50, 100, 500, and 1000 were sampled from each of the generators and MW U tests were conducted on each of these synthetic datasets separately. The DP-hyperparameters were all set to C = 1 for the gradient clipping bound and 1.07 for the noise multiplier as in \cite{chen2020gs}. 

A summary of the settings for the experiments with original Gaussian data is provided in Table \ref{table_1}.

\begin{table}[htbp]
\centering
\begin{adjustbox}{width=1\textwidth}
\begin{tabular}{|l|l|l|l|}
\hline \hline
\textbf{DP Method}     & \multicolumn{1}{c|}{\textbf{\begin{tabular}[c]{@{}c@{}}Original Dataset Size\\ Group ratio 50\%\end{tabular}}} & \multicolumn{1}{c|}{\textbf{\begin{tabular}[c]{@{}c@{}}DP-synthetic \\ Dataset Size\end{tabular}}}  & \textbf{Privacy Budget}                        \\ \hline
DP-MW U test           & 50, 100, 500, 1000, 20 000                                                                                     & N/A                                             & $\epsilon = 0.01, 0.1, 1, 5, 10$                  \\ \hline
DP Perturbed Histogram & \multirow{3}{*}{50, 100, 500, 1000, 20 000}                                                                    & \multirow{3}{*}{\begin{tabular}[c]{@{}l@{}}Similar to the \\ original dataset\end{tabular}} & \multirow{3}{*}{$\epsilon = 0.01, 0.1, 1, 5, 10$} \\
Private-PGM            &                                                                                                                &                                                  &                                                \\
MWEM                   &                                                                                                                &                                                  &                                                \\ \hline
DP Smoothed Histogram    & 20 000                                                                                                         & 50, 100, 500, 1000                               & $\epsilon = 0.01, 0.1, 1, 5, 10$                  \\ \hline
DP GAN                 & 20 000                                                                                                         & 50, 100, 500, 1000                               & $\epsilon = 1, 2, 3, 4, 5, 10$                    \\ \hline \hline
\end{tabular}
\end{adjustbox}
\caption{\label{table_1}Setup for experiments using original Gaussian data. For the DP-MW U test, DP-synthetic dataset size is not applicable ('N/A'), because this method is computed on the original sensitive data.}
\end{table}

\subsection*{\textbf{\small Setup for Real-World Data}}
The size of the prostate cancer dataset constrained some of the experiments. Therefore, DP Smoothed Histogram and DP GAN experiments with this dataset were excluded, as these methods require a larger original dataset size (i.e., thousands of observations) to apply them accurately. On the other hand, the cardiovascular dataset size allowed us to carry out experiments with all the DP methods. 

In the experiments with the prostate cancer dataset, we applied each considered DP method at each epsilon value 1000 times. While in the cardiovascular dataset experiments, we used the data to sample 1000 original datasets for each dataset size 50, 100, 500, 1000, and 20 000; then, for each sampled dataset, we applied the DP methods at each epsilon. The proportion of Type II error was measured over the 1000 repetitions for each experiment setting. For DP Smoothed Histogram and DP GAN, due to their nature, the experiments were performed differently; however, they had a similar setting to the ones with Gaussian signal data.

\subsection*{\textbf{\small Setup for Simulated Multivariate Data}}
In the experiments with a simulated multivariate dataset, we considered the Private-PGM and MWEM synthesizers. Using the generated DP-synthetic data, we empirically assessed the proportion of Type I and Type II errors for the MW U test for an ordinal variable (PI-RADS score), Student’s t-test for a normally distributed continuous variable (age), median test for another continuous variable (PSA), and chi-squared test of independence for a binary variable (use of 5ARIs medication).

For these experiments, we generated 1000 simulated multivariate datasets for dataset sizes of 50, 100, 500, 1000, and 20000. Subsequently, for each simulated dataset, we applied the DP-synthetic data generator at each epsilon value. The proportion of Type I and Type II error were measured across the DP-synthetic datasets, with the condition that the 
requirements for running the statistical test were met in at least 50 of the generated DP-synthetic datasets (see supplementary material A.2 for further details on cases when tests are undefined, such as when the DP-synthetic data consists of only single class).

\section{Results }

\subsection{Gaussian Data}
\begin{figure}[htpb]
    \centering
    \includegraphics[width=0.99\textwidth]{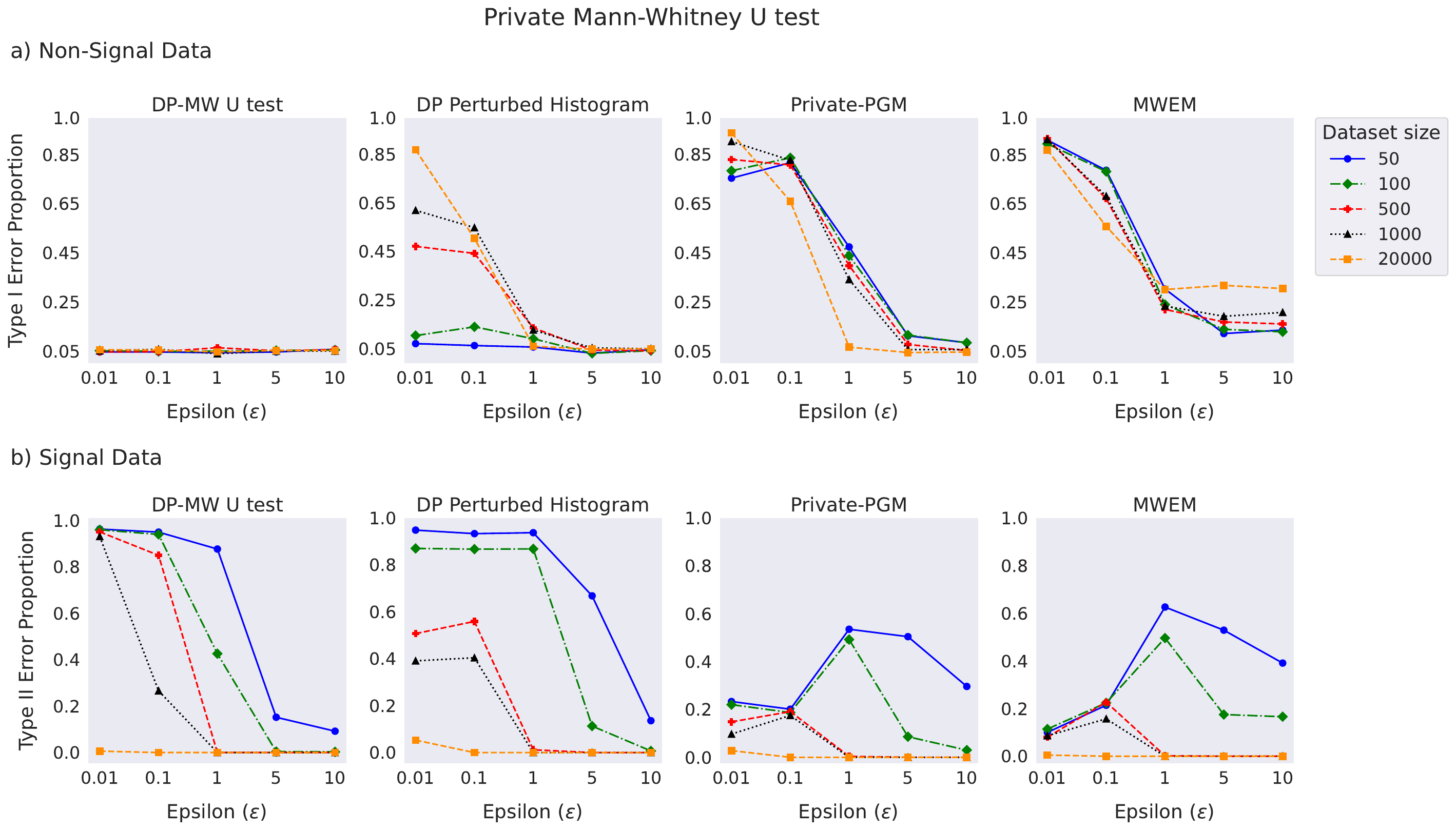}   
    \caption{\small The proportion of Type I and Type II errors for the Mann-Whitney U test using four differentially private (DP) methods: DP-MW U test, DP Perturbed Histogram, Private-PGM, and MWEM at different privacy budget ($\epsilon$). The dataset size indicates the size of the original data used in the experiments by the DP methods. The proportions of Type I error and Type II error were measured over 1000 repetitions of the experiment using Gaussian a) non-signal data and b) signal data, respectively.}
    \label{fig:figure_4}
\end{figure}

In Figure \ref{fig:figure_4}(a), experiments on  Gaussian non-signal data (i.e., both groups originate from the same Gaussian distribution) show that when the DP-MW U test is applied to the 1000 datasets, the proportion of Type I stays close to $\alpha = 0.05$ for all dataset sizes at all $\epsilon$. Meanwhile, the MW U test on DP-synthetic data from DP Perturbed Histogram, Private-PGM, and MWEM have a high proportion of Type I error for $\epsilon < 5$, falsely indicating a significant difference between the two groups. From these DP methods, DP Perturbed Histogram and Private-PGM benefit of having a large original dataset size (i.e., 20 000), as $\epsilon$ can be reduced to 1 while still having a Type I error close to $\alpha = 0.05$. MWEM is the method with the worst performance as the proportion of Type I error for all sample sizes stays above 0.05 even for $\epsilon = 10$. 

Figure \ref{fig:figure_4}(b) presents the results for Gaussian signal data where a difference between the two groups exists (i.e., normally distributed data of two groups with means one standard deviation apart). From these results, we observed that the MW U test Type II error for all the DP methods, with low $\epsilon$, can be reduced by increasing the dataset size, corroborating the trade-off that exists between privacy, utility, and dataset size.

\begin{figure}[ht]
    \centering
    \includegraphics[width=0.99\textwidth]{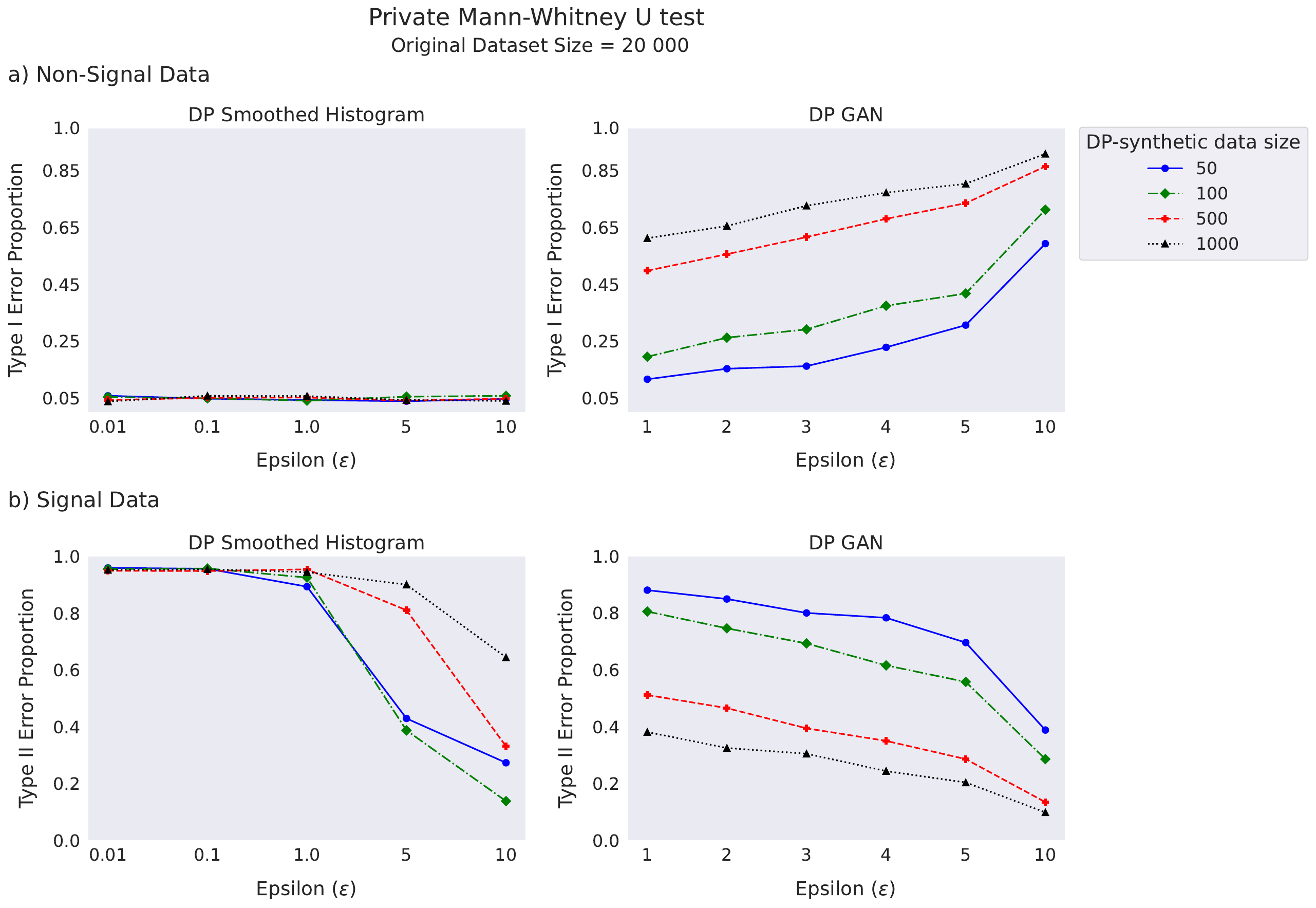}
     \caption{\small The proportion of Type I and Type II error of MW U test applied to synthetic data generated from DP Smoothed Histogram and DP GAN. The size of the original dataset is 20 000 with a group ratio of 50\%. DP-synthetic data of sizes 50, 100, 500, and 1000 were generated from both methods. The proportions of Type I error and Type II error were measured over 1000 DP-synthetic datasets using Gaussian a) non-signal data and b) signal data, respectively.}
   \label{fig:figure_5}
\end{figure}

Results for the MW U test on DP-synthetic data from DP Smoothed Histogram and DP GAN are presented in Figure \ref{fig:figure_5}. The DP Smoothed Histogram method controls the  Type I error reliably. However, the price for this is that in most of our experiment settings, it has high Type II error, meaning that the real difference between the groups present in the original data is lost in the DP-synthetic data generation process. DP GAN shows very high Type I error that as an interesting contrast to the other methods grows as privacy level is reduced.

\begin{figure}[htpb]
    \centering
    \includegraphics[width=0.99\textwidth]{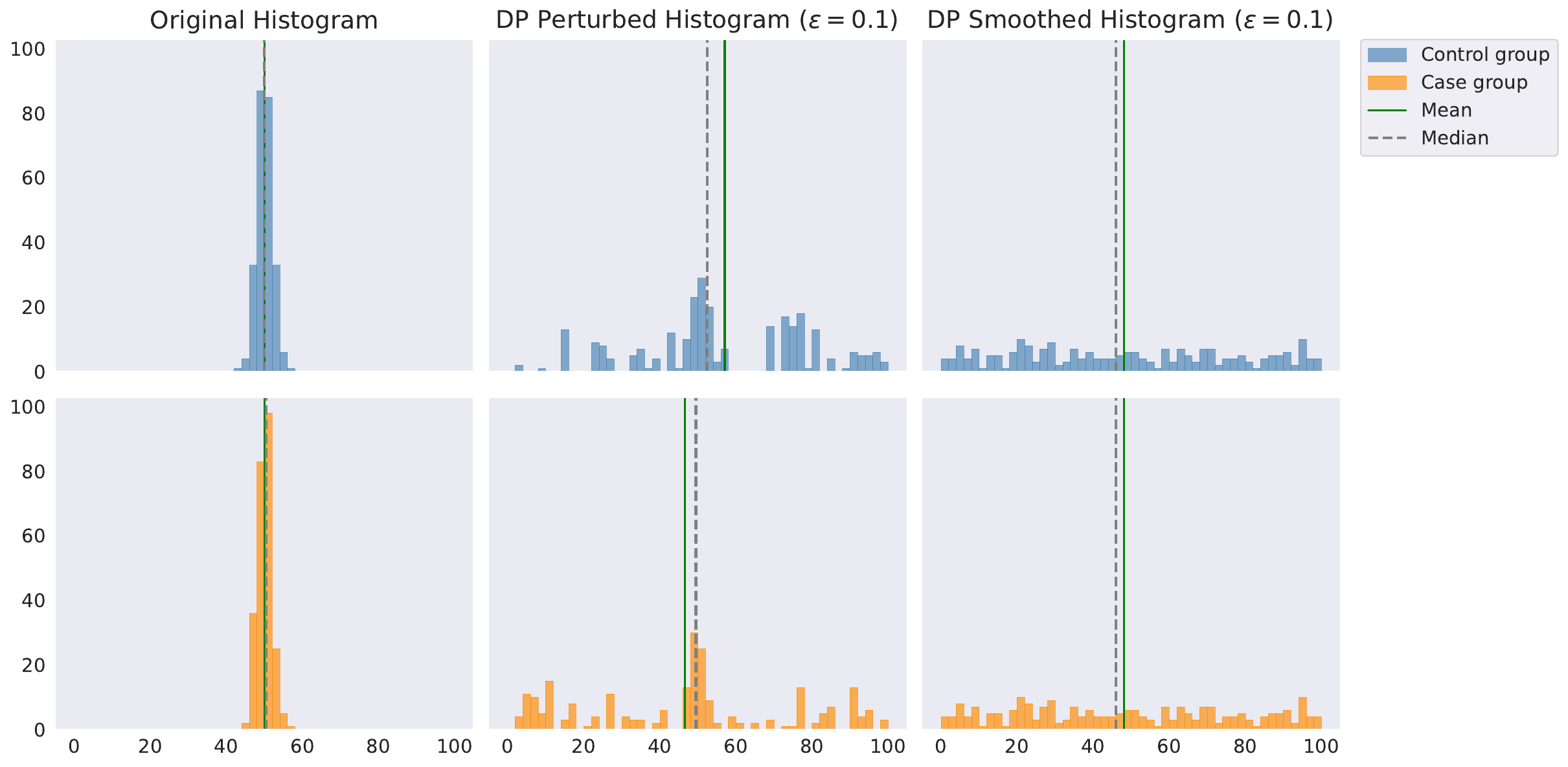}
     \caption{\small Example of the two groups distributions in a non-signal original dataset of size 500 (U stat = 31460.5 p-value = 0.8953) and corresponding distributions for synthetic data generated using DP Perturbed Histogram (MW U stat = 38191.5 p-value = 0.00001774) and DP Smoothed Histogram (MW U stat = 29621.5 p-value = 0.3314) with $\varepsilon=0.1$ as the privacy budget. With such high level of privacy enforced neither of the DP-synthetic datasets preserve well the structure of the original data. However, the DP Perturbed Histogram has the tendency to create artificial differences between the two groups such that result in low p-values for MW U test, whereas with DP Smoothed Histogram method both the generated case and control groups follow similar close to uniform distributions.}
   \label{fig:figure_6}
\end{figure}

To summarize, these results show that except for DP Smoothed Histogram, all the DP-synthetic data generation methods have highly inflated Type I error. This means that they are prone to generating data from which false discoveries are likely to be made. For the histogram-based methods, increased Type I error was associated with increased level of privacy, the effect being especially clear for $\epsilon < 5$. Figure \ref{fig:figure_6} presents an example of false discovery on synthetic data generated with the DP Perturbed histogram at $\epsilon = 0.1$, and also demonstrates how DP Smoothed Histogram does not exhibit the same behavior.

\subsection{Real-world Data}
\begin{figure}[htbp]
    \centering
    \includegraphics[width=0.99\textwidth]{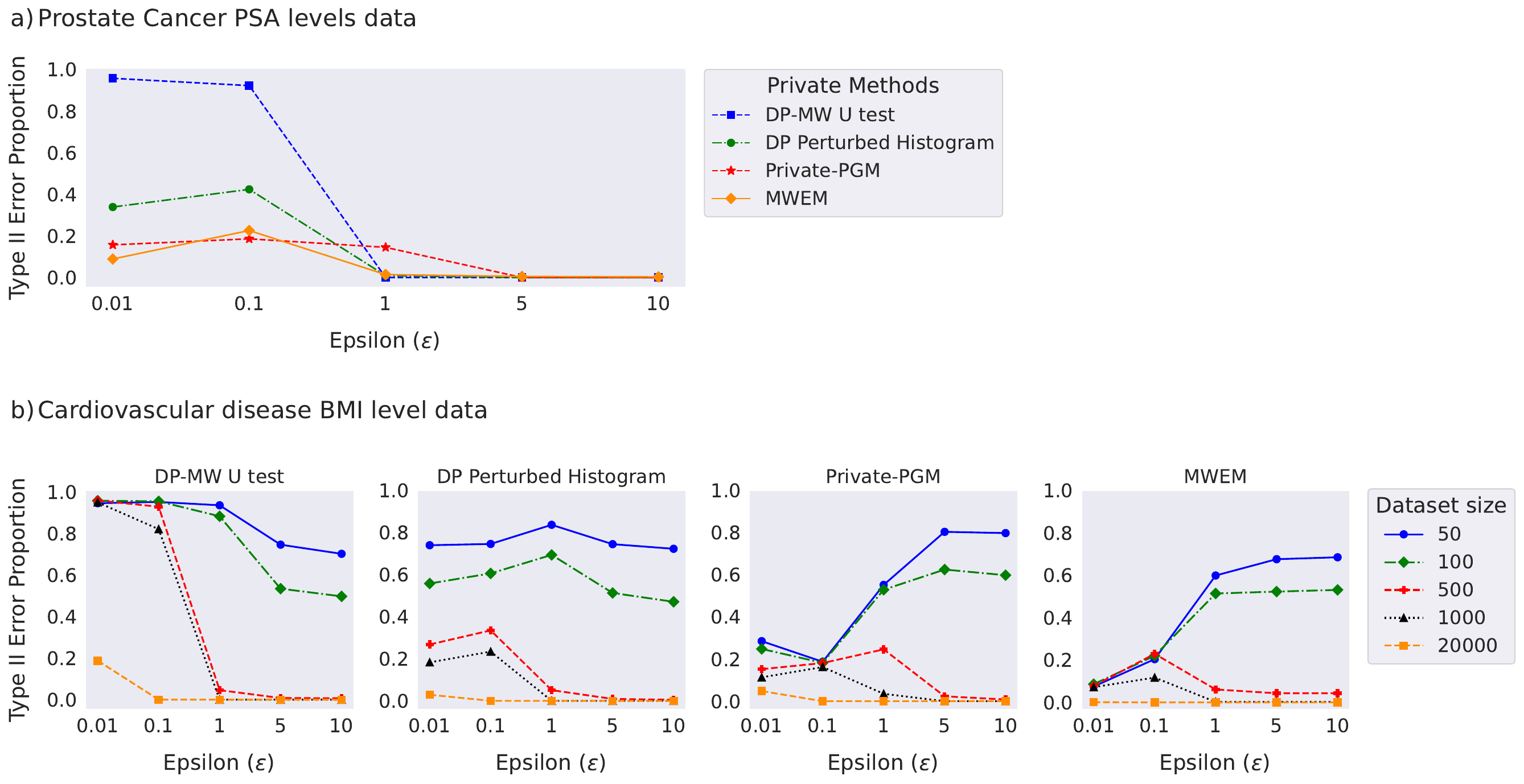}   
    \caption{\small The proportion of Type II error for the Mann-Whitney U test using four DP methods: DP-MW U test, DP Perturbed Histogram, Private-PGM and MWEM applied to a) the PSA level data in the prostate cancer dataset (dataset size = 500), b) the body mass index (BMI) data in the Kaggle cardiovascular disease dataset.}
    \label{fig:figure_7}    
\end{figure}

 Figure \ref{fig:figure_7}(a) shows the results of experiments conducted with the prostate cancer dataset. The DP-MW U test performs as expected for an original dataset size of 500 with a group ratio of approx. 50\%. The null hypothesis is rejected for $\epsilon \geq 1$, while for $\epsilon < 1$ it is often not rejected. Similar behavior is present in DP-synthetic data from DP Perturbed Histogram and MWEM, yet the chance of rejecting the null hypothesis when $\epsilon < 1$ is higher than in the DP-MW U test. In DP-synthetic data from Private-PGM, the null hypothesis is rejected for $\epsilon \geq 5$ more often than for $\epsilon < 5$.
 
\begin{figure}[htbp]
 \centering
  \includegraphics[width=0.95\textwidth]{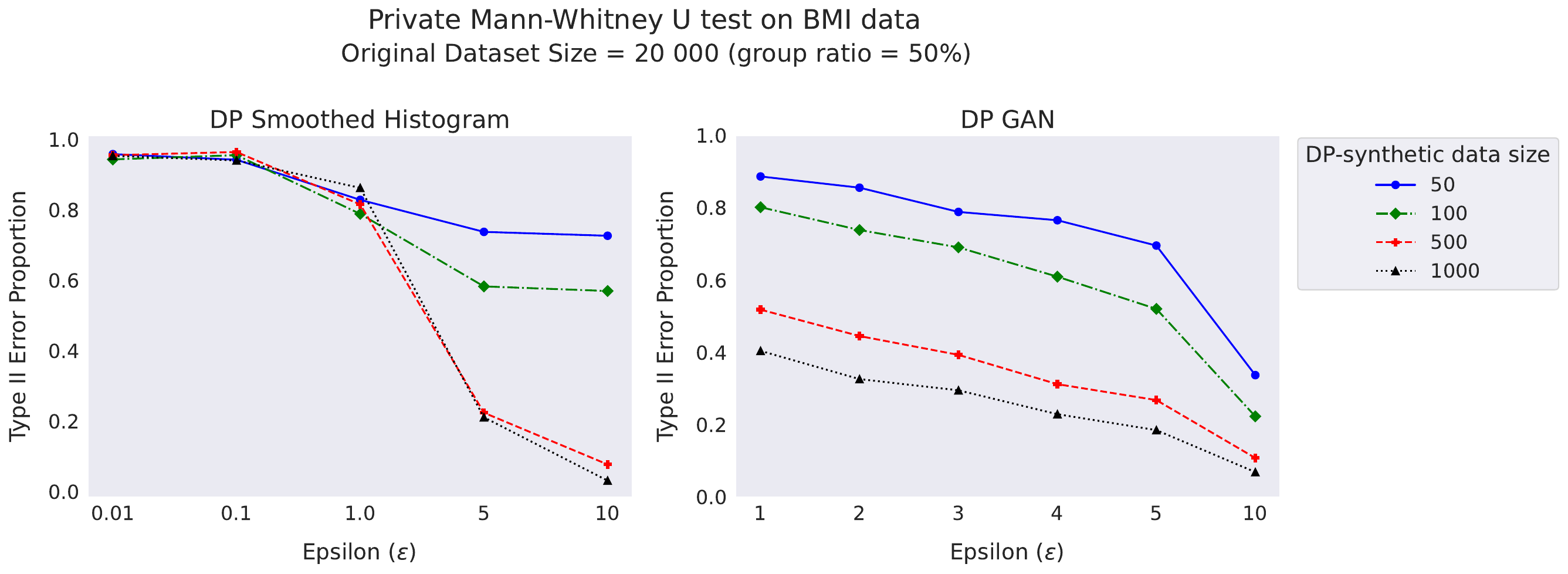}
   \caption{\small The proportion of Type II error of MW U test applied to synthetic data generated with DP Smoothed Histogram and DP GAN. The original dataset of size 20 000 with a group ratio of 50\% was drawn from the publicly available Cardiovascular disease dataset. DP-synthetic data of sizes 50, 100, 500, and 1000 were generated using both DP methods 1000 times.}
 \label{fig:figure_8}
\end{figure}

The experiment results for the DP-MW U test, DP Perturbed Histogram, Private-PGM, and MWEM applied to the Cardiovascular disease dataset are presented in Figure \ref{fig:figure_7}(b). In this dataset, we observe that MWEM and Private-PGM are the methods that benefit the most from increasing the original sample size, as stronger privacy guarantees can be provided without the MW U test losing power. These results agree with the ones obtained when using Gaussian signal data.

Results for DP Smoothed Histogram and DP GAN applied to the cardiovascular dataset are presented in Figure \ref{fig:figure_8}. With DP Smoothed Histogram, Type II error is on an acceptable level when $\epsilon\geq5$ and the sample size is 500 or 1000, whereas for lower $\epsilon$ values the effect is not found. DP GAN results have lower Type II error, but given how high Type I error the method shows in the non-signal experiments the approach is less reliable compared to the DP Smoothed Histogram method.

\subsection{Simulated Multivariate Data}

\begin{figure}[htbp]
 \centering
  \includegraphics[width=0.90\textwidth]{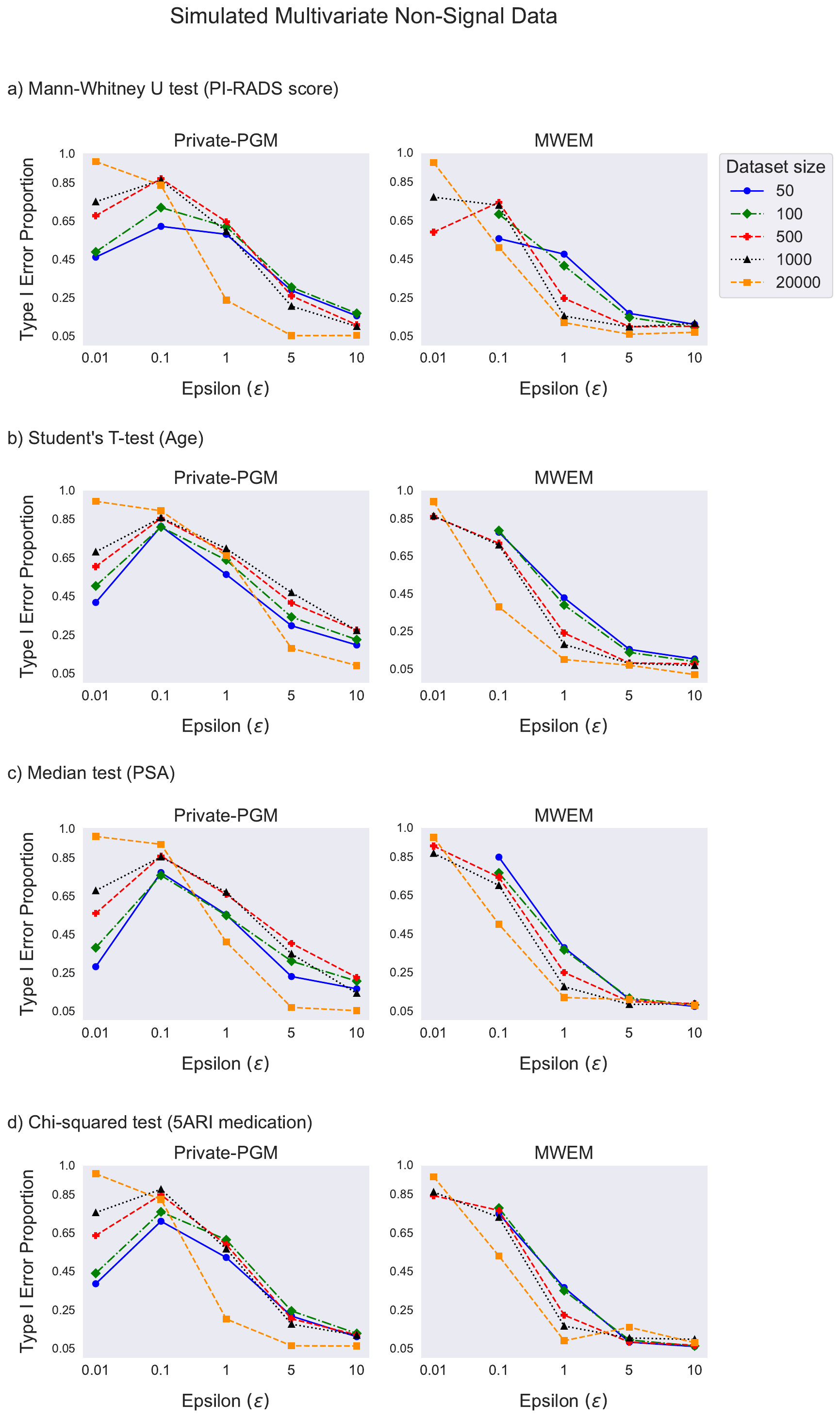}
   \caption{\small Proportion of Type I error conditioned to the number of DP-synthetic datasets where the statistical test is feasible; a) MW U test applied to an ordinal variable (PI-RADS score); b) Student’s T-test on a normally distributed variable (Age); c) Median test on continuous variable (PSA); d) Chi-squared test on a binary variable (5ARI medication).}
 \label{fig:figure_9} 
\end{figure}

\begin{figure}[htbp]
 \centering
  \includegraphics[width=0.90\textwidth]{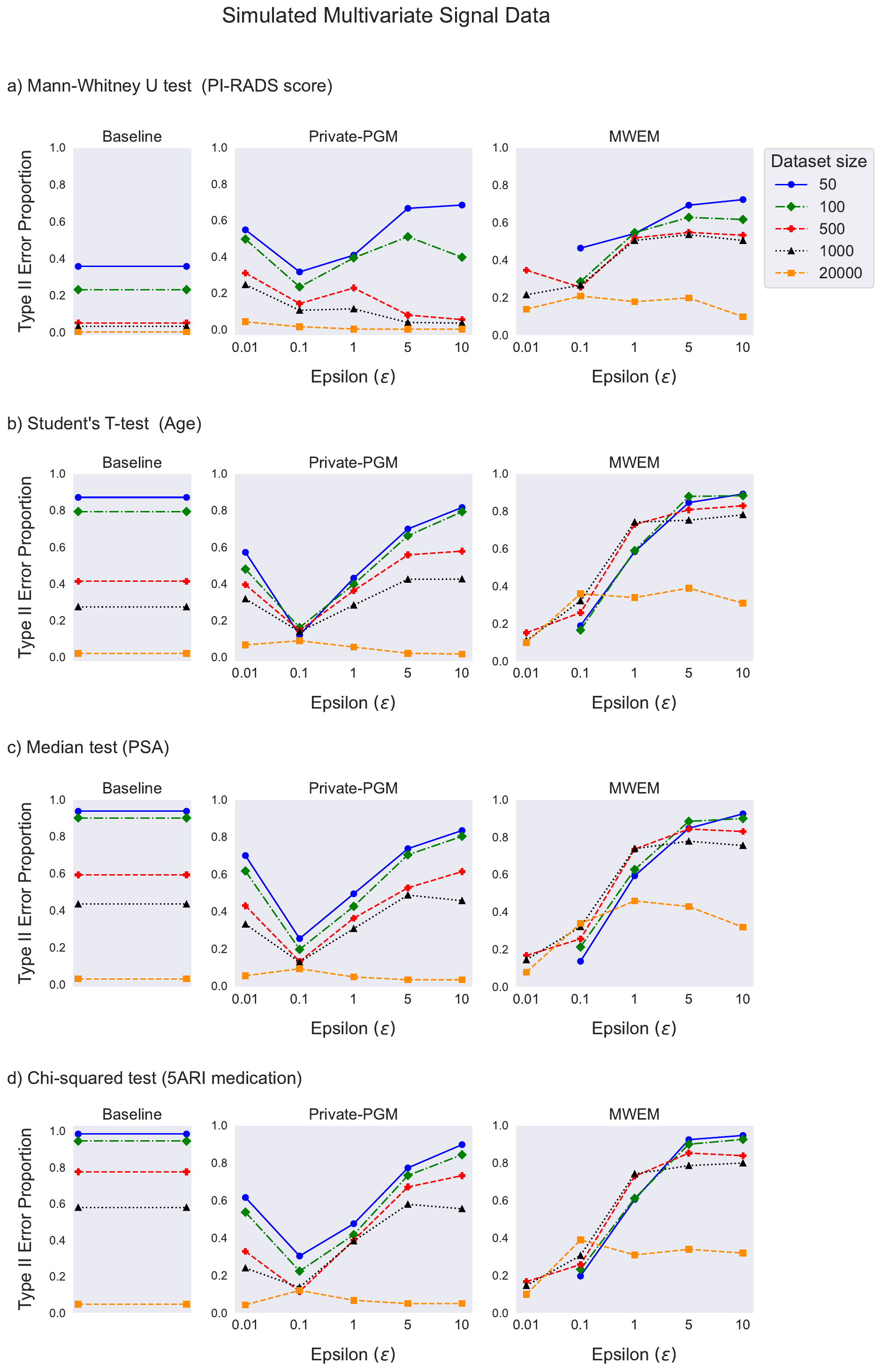}
   \caption{\small Proportion of Type II error conditioned to the number of DP-synthetic datasets where the statistical test is feasible; a) MW U test applied to an ordinal variable (PI-RADS score); b) Student’s T-test on a normally distributed variable (Age); c) Median test on continuous variable (PSA); d) Chi-squared test on a binary variable (5ARI medication).}
 \label{fig:figure_10} 
\end{figure}

In Figure \ref{fig:figure_9}, the proportion of Type I errors for various statistical tests (i.e., MW U test, Student’s t-test, median test, and chi-squared test) is presented. From these results, we observe that false discoveries are also prone to occur similarly to the previous experiments with only two variables. The validity of the tests is preserved only for largest tested privacy budgets combined with large amounts of the original sensitive data.
Same kind of trend was observed for all statistical tests under consideration. For Private-PGM, a substantial drop in Type I error was observed for $\epsilon=0.01$ and dataset size $ < 20 000$. On closer examination, we observed that with the smallest privacy budgets the size of the smaller of the two groups tends to be very small or even zero. This can be seen from the numbers of times the test requirements failed, as presented in supplementary
material A.2, where the tests fail when the size of smaller group is zero. The power of all evaluated tests strongly depend on the group size imbalance in the sample, so that for a fixed sample size they have the highest power for equal group sizes and the power shrinks to zero when the smaller group size goes to zero. Therefore, the tendency of the low privacy budgets to produce imbalanced samples counters the tendency to produce fake group differences to some extent.

In the case of Type II error proportions (Figure \ref{fig:figure_10}), the results depend on the magnitude of the group differences in the original data, how it is preserved by the GaussianCopulaSynthesizer, and the size of the simulated dataset. As a baseline or point of reference, we first present the Type II error probabilities computed over the 1000 simulated multivariate datasets that represent the original sensitive data before the synthetic data is generated based on them. Then, we illustrate the corresponding Type II error probabilities for the DP-synthetic data with different privacy budgets.
For the synthetic data, especially for $\epsilon<10$, we observe that the Type II errors are often lower than those of on the original data, indicating that true group differences are discovered more often from the synthetic data than from the original.
However, this is explained perfectly by the large Type I error probabilities presented in Figure~\ref{fig:figure_9}, indicating that the fake group differences present in the synthetic data are so strong that they end up getting discovered rather than the true ones that are too weak to be discovered from the original data. For $\epsilon=0.01$, the large Type II error of Private-PGM also mirrors the low Type I error, caused by the loss of power due to the group size imbalance.

\section{Discussion}
This study investigated to what extent the validity and power of independent sample tests are preserved in DP-synthetic data. Experimental results on Gaussian, real-world, and multivariate simulated data demonstrate that the generated DP-synthetic data, especially with strong privacy guarantees ($\epsilon \leq 1$), can lead to false discoveries. We empirically show that many state-of-the-art DP methods for generating synthetic data have highly inflated Type I error rates when the privacy level is high. These results indicate that false discoveries or inferences are likely to be drawn from the DP-synthetic data produced by these DP methods. Our findings are in line with other studies that have presented or stated that DP-synthetic data can be invalid for statistical inference and indicated the need for methods that are noise-aware in order to produce accurate statistical inferences \cite{gaboardi2016differentially, charest2011can, charest2012empirical, giles2022faking, raisa2023noise}. 

Additionally, it is necessary to be cautious when analyzing Type II error results, as this is only meaningful for valid tests where the Type I error is properly controlled. The Type II error tends to decrease with the increase of Type I error, as these errors are inversely related. In our study, the only DP method based on synthetic data generation that had a valid Type I error over all the privacy budgets tested was the DP Smooth Histogram method. However, the method is applicable only when the original dataset size is fairly large (e.g. n= 20 000 in our experiments), and tended to have high Type II error when the amount of privacy enforced was high (e.g. $\epsilon\leq1$). For DP Perturbed Histogram and Private-PGM methods both Type I and Type II errors remained low for $\epsilon\geq5$, whereas MWEM and DP GAN did not provide valid Type I error levels even with lowest privacy values tested.

The main advantage of releasing DP-synthetic data, as opposed to releasing only analysis results from the original data,  is that it can be ideally used to support a wide range of analyses by different users. Due to post-processing property of DP, any type or number of analyses done on the synthetic data are also guaranteed to be DP with no further privacy budget needed. However, if the only goal is to perform a limited number of pre-defined analyses, it makes more sense to do these on the original data with DP methods. This is illustrated in our experiments by the DP-MW U test baseline that always outperforms analyses done on DP-synthetic data. As a middle ground between these approaches, an active area of research is to develop such  DP synthetization methods where the data is optimized to support certain types of analyses well, such as PrivPfC for classifier training \cite{su2018privpfc} and various Bayesian noise-aware DP synthetic data generation methods \cite{raisa2023noise}.

There are limitations in our study that could be addressed in future research. One limitation is that the marginals or histograms based DP methods require continuous variables to be discretized. This discretization must be performed in a private matter or based on literature to avoid leaking private information. Besides, it is well-known that the number of bins used to discretize the data has a significant impact on the quality of the resulting data \cite{xu2013differentially, mckenna2022aim}. Therefore, choosing the number of bins is problem- and data-dependent and can affect the results. In our experiments with Gaussian data, the continuous values were discretized using 100 bins. This number of bins was selected to show a possible extreme case where having bins empty or with small counts deteriorates the quality of the generated DP-synthetic data. On the other hand, for our experiments with real-world and multivariate simulated data, the number of bins used was determined based on domain knowledge and literature. Finally, testing different hyperparameter values for the DP method implementations could yield different results for the methods. 

\section{Conclusions}
Our results suggest caution when releasing DP-synthetic data, as false discoveries or loss of information are likely to happen especially when a high level of privacy is enforced. To an extent, these issues may be mitigated by having large enough original datasets, selecting methods that are less prone to adding false signal to data, and by carefully comparing the quality of the DP-synthetic data to the original one based on various quality metrics (see e.g. \cite{hernadez2023synthetic}) before data release. Still, with current methods DP-synthetic data may be a poor substitute for real data when performing statistical hypothesis testing, as one cannot be sure if the results obtained are based on trends that hold true in the real data, or due to artefacts introduced when synthesizing the data.

\section{Acknowledgements }
This work has received funding from Business Finland (grant number 37428/31/2020) and European Union’s Horizon Europe research and innovation programme (grant number 101095384). Views and opinions expressed are however those of the authors only and do not necessarily reflect those of the European Union or the European Health and Digital Executive Agency (HADEA). Neither the European Union nor the granting authority can be held responsible for them. The authors would like to express their gratitude to Peter J. Boström, Ivan Jambor, and collaborators for their support and contribution in providing the prostate cancer datasets used in the real-world data experiments. We also thank Katariina Perkonoja for her insightful feedback regarding the experimental setup and the statistical tests, as well as the anonymous reviewers for their valuable comments.

\bibliographystyle{ieeetr}
\bibliography{sample}

\clearpage

\appendix
\section{Supplementary Material}
\subsection{Drawing data from additively smoothed histograms is
differentially private}\label{appendix:proof}

\newcommand{\countvec}{c}
\newcommand{\histlen}{h}
\newcommand{\scorevec}{s}
\newcommand{\synthsize}{m}
\newcommand{\probvec}{p}
\newcommand{\resvec}{r}

\setcounter{table}{0}
\makeatletter 
\renewcommand{\thetable}{S\arabic{table}}

Proof: 

Drawing a single datum from a distribution determined by a histogram can be carried out in a differentially private way using the  exponential mechanism, in which the original probabilities are modified as follows. Instead of the relative size of a histogram bin, each possible outcome of the draw has its probability proportional to \begin{align}\label{expmech}
e^{\frac{\epsilon\scorevec_i}{2\Delta}}\;,
\end{align}
where $\scorevec_i$ is the score of the $i$th option (e.g. histogram bin) and $\Delta$ is the sensitivity of the scoring, indicating the maximum amount the score can change for any outcome if a single datum is changed among the sensitive data. By substituting (\ref{expmech}) into (\ref{logprobs}), it is easy to verify that the exponential mechanism fulfills differential privacy. If we define the score for the $i$th output as $s_i=\alpha\ln (\countvec_i + \alpha)$, where $\countvec_i$ is the number of sensitive data in the $i$th bin and $\alpha=\frac{2}{\epsilon}$, the maximal change of the score value takes place when $\countvec_i=1$ changes to $\countvec_i=0$ or vice versa. For this maximal change, we have the following upper bound:\begin{align*}
\alpha\ln (1 + \alpha)-\alpha\ln (\alpha)
&=\alpha\ln\left(1 + \frac{1}{\alpha}\right)\\
&=\ln\left(\left(1 + \frac{1}{\alpha}\right)^\alpha\right)\\
&<\ln(e)\\
&=1\;,
\end{align*}
where the inequality follows from the well-known property of $e$ (see e.g. pages 44-48 of \cite{dorie1965}):
\[
\left(1 + \frac{1}{x}\right)^x<e,
\]
for all $x>0$. Thus, the sensitivity of the scoring is 1. Substituting the scoring into (\ref{expmech}) indicates that drawing a single datum according to the probabilities proportional to $\countvec_i+\frac{2}{\epsilon}$ is $\epsilon$ differentially private.

Due to the sequential composability property, drawing $\synthsize$ synthetic data instead of only one datum requires $\synthsize$ times larger privacy budget. Consequently, drawing $\synthsize$ according to the probabilities proportional to $\countvec_i+\frac{2\synthsize}{\epsilon}$ is $\epsilon$ differentially private.

\begin{algorithm}[H]
	\caption{Draw synthetic sample}\label{alg:cap}
	\begin{algorithmic}
		\Require $\bm{\countvec}$ \Comment{Histogram count vector of length $\histlen$}
		\Require $\synthsize$ \Comment{Number of synthetic data to be drawn}
		\Require $\epsilon$
		%\Ensure 
		\State $\bm{\scorevec} \gets$ score vector of length $\histlen$
		\State $\bm{\probvec} \gets$ probability vector of length $\histlen$
		\State $\bm{\resvec} \gets$ synthetic data count vector of length $\histlen$
		\State $\alpha \gets 2 \synthsize / \epsilon$ \Comment{Calculate amount of additive smoothing}
		\For{$i \in \{1,\ldots,\histlen\}$}
		\State $\bm{s}_i\gets \alpha\ln (\countvec_i + \alpha)$\Comment{Calculate scores for bins}
		\EndFor
		\For{$i \in \{1,\ldots,\histlen\}$}
		\State $\probvec_i\gets e^\frac{\epsilon\scorevec_i}{2\synthsize}$\Comment{Calculate probabilities according to the exponential mechanism}
		\EndFor
		\For{$i \in \{1,\ldots,\histlen\}$}
		\State $\probvec_i\gets\probvec_i/\Arrowvert\bm{\probvec}\Arrowvert_1$\Comment{Normalize the probabilities to sum up to one}
		\EndFor
		\State $\bm{\resvec}\gets$ Draw $\synthsize$ synthetic data according to the probabilities $\bm{\probvec}$.
		\State \Return $\bm{\resvec}$
	\end{algorithmic}
\end{algorithm}

\subsection{Failed DP-synthetic data sets}

DP-synthetic data generation methods can sometimes produce data sets that fail to meet the requirements of a statistical test, even if the original sensitive data would fulfill these. This type of degenerate behavior was observed in the experiments especially when generating multivariate data for very small $\epsilon$ values and small dataset sizes. Typical failure mode observed was that only one class was represented in the synthetic data, in which case none of the tests apply. Further, in some cases all the values of continuous variable were observed to be the same in the synthetic data, in which case t-test is undefined. In the case of the chi-squared test applied to binary variables, highly imbalanced or sparse synthetic data resulted in low expected frequencies, meaning that one or more cells in the contingency table had an expected frequency of less than five, causing the test to fail. In the experiments, Type I and Type II was calculated based on only those generated data sets that met the requirements of the considered tests. Tables \ref{table:s1} and \ref{table:s2} catalogue the number of failed data sets for the experiments performed on the multivariate simulated prostate cancer data.

 \begin{table}[htbp]
    \centering
    \includegraphics[width=0.99\textwidth]{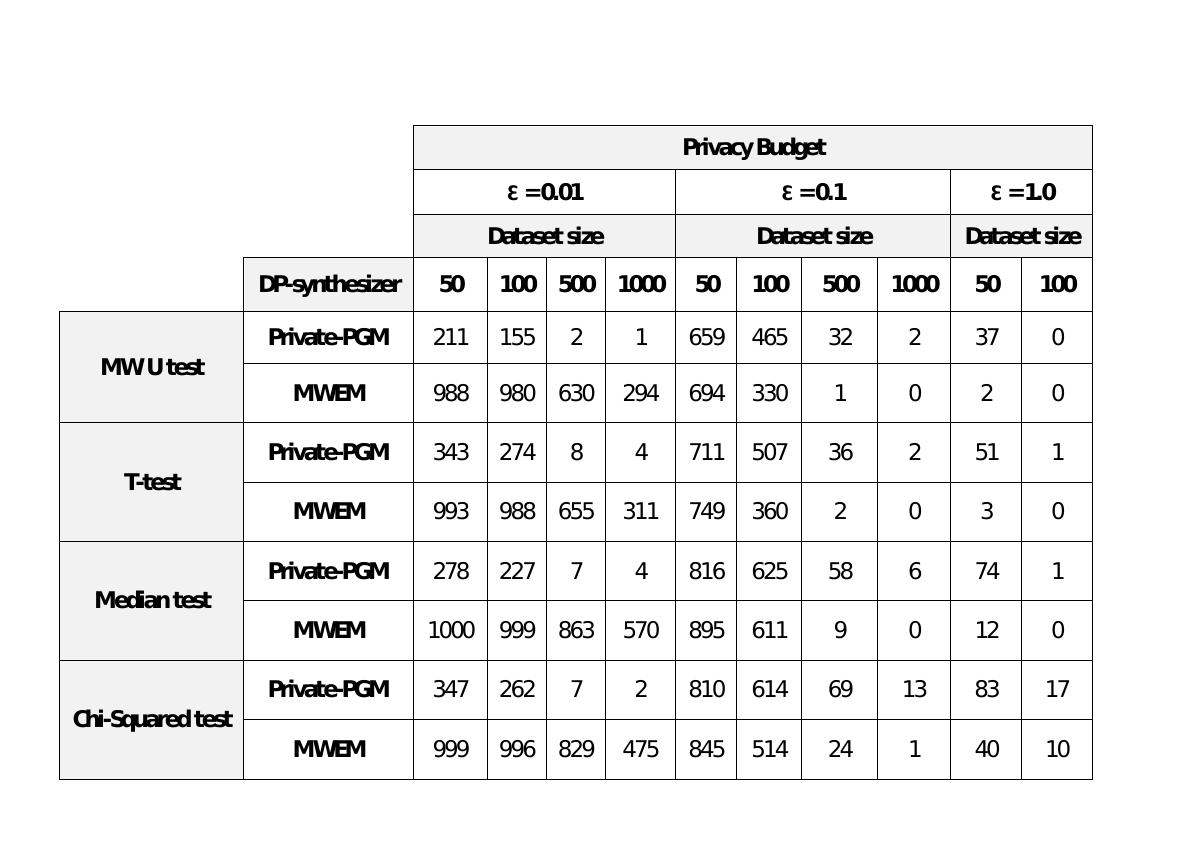}   
    \caption{\label{Table_S1}\small Number DP-synthetic datasets that failed to meet the requirements of the statistical test out of 1000 generated for each corresponding DP-synthesizer, privacy budget, and dataset size on simulated multivariate non-signal data. There were no failures in experiments with sample sizes of 20 000, as well as in datasets with sizes of 50, 100, 500, and 100 and $\varepsilon \geq 5$. }
    \label{table:s1} 
\end{table}

\begin{table}[htbp]
    \centering
    \includegraphics[width=0.99\textwidth]{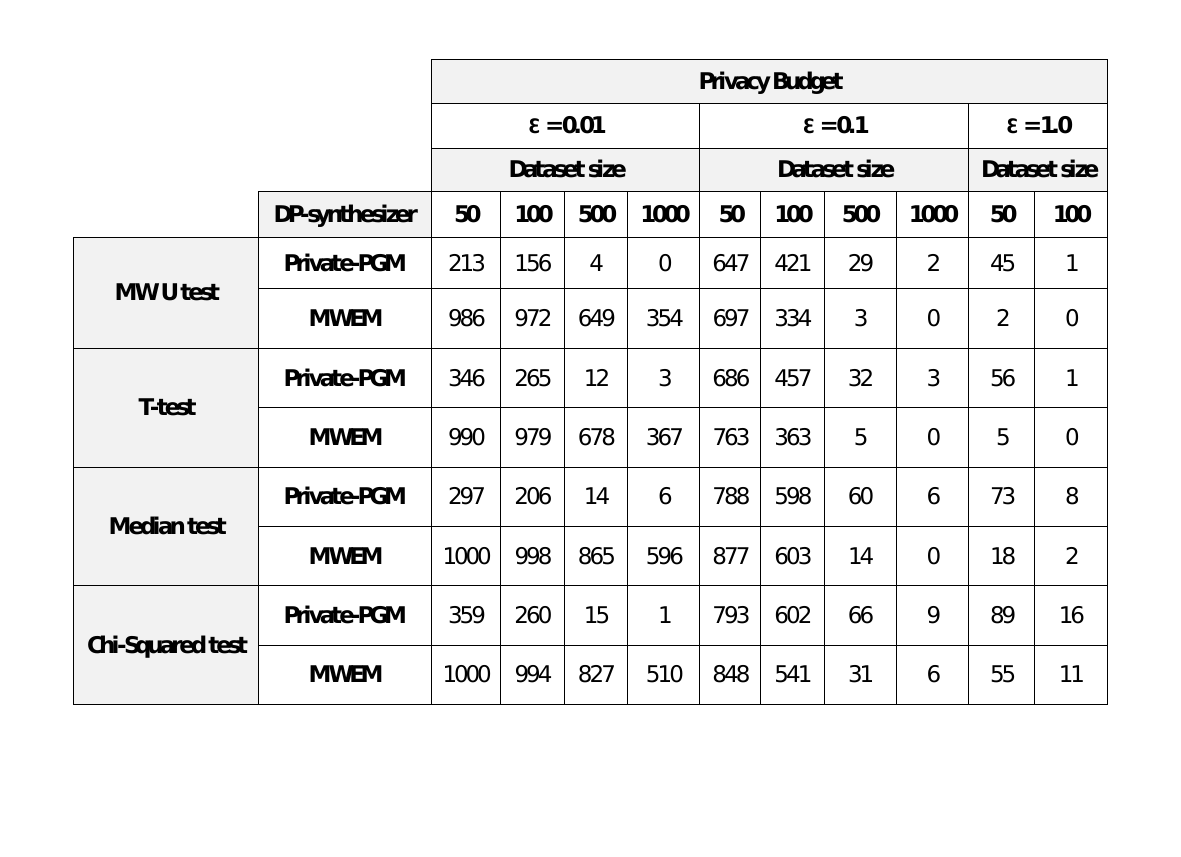}   
    \caption{\label{Table_S2}\small Number DP-synthetic datasets that failed to meet the requirements of the statistical test out of 1000 generated for each corresponding DP-synthesizer, privacy budget, and dataset size on simulated multivariate signal data. There were no failures in experiments with sample sizes of 20 000, as well as in datasets with sizes of 50, 100, 500, and 100 and $\varepsilon \geq 5$.}
    \label{table:s2} 
\end{table}

\end{document}